\pdfoutput=1

\documentclass[11pt]{article}

\usepackage[final]{acl}

\usepackage{times}
\usepackage{latexsym}
\usepackage{url}

\usepackage[T1]{fontenc}

\usepackage[utf8]{inputenc}

\usepackage{microtype}

\usepackage{inconsolata}

\usepackage{graphicx}

\usepackage{comment}
\usepackage{caption}
\usepackage{subcaption}
\usepackage{algorithm2e}

\usepackage{todonotes}

\title{\emph{Am I eligible?} Natural Language Inference\\ for Clinical Trial Patient Recruitment: the Patient's Point of View}

\author{Mathilde Aguiar, Pierre Zweigenbaum, Nona Naderi  \\
         Université Paris-Saclay, CNRS, Laboratoire Interdisciplinaire des Sciences du Numérique, \\
         91405, Orsay, France \\
         \texttt{first.last@lisn.fr} \\}

\begin{document}
\maketitle
\begin{abstract}
Recruiting patients to participate in clinical trials can be challenging and time-consuming. 
Usually, participation in a clinical trial is initiated by a healthcare professional and proposed to the patient. Promoting clinical trials directly to patients via online recruitment might help to reach them more efficiently.
In this study, we address the case where a patient is initiating their own recruitment process and wants to determine whether they are eligible for a given clinical trial, using their own language to describe their medical profile.
To study whether this creates difficulties in the patient-trial matching process,
we design a new dataset and task, Natural Language Inference for Patient Recruitment (NLI4PR), in which patient-language profiles must be matched to clinical trials.
We create it by adapting the TREC 2022 Clinical Trial Track dataset, which provides patients' medical profiles, and rephrasing them manually using patient language. We also use the associated clinical trial reports where the patients are either eligible or excluded. 
We prompt several open-source Large Language Models 
on our task and achieve from 56.5 to 71.8 of F1 score using patient language, against 64.7 to 73.1 for the same task using medical language.
When using patient language, we observe only a small loss in performance for the best model, suggesting that having the patient as a starting point could be adopted to help recruit patients for clinical trials. 
The corpus and code bases are all freely available on our Github\footnote{\url{https://github.com/CTInfer/NLI4PR}} and HuggingFace\footnote{\url{https://huggingface.co/datasets/Mathilde/NLI4PR}} repositories.
\end{abstract}

\section{Introduction}

Many efforts have been made to develop methods based on Natural Language Processing (NLP) to solve ongoing challenges in healthcare. These studies are targeting either medical professionals or patients.
However, patients and medical professionals use different kinds of language. 
A system trained and designed on medical language might, therefore, fail when used with patient language. 
Before releasing a new medicine on the market,
clinical trials must be performed and
recruit several cohorts of patients
with
profiles that comply with the inclusion and exclusion criteria of the trial. Recruiting patients
can be challenging and costly, especially for studies focusing on certain diseases or
targeting
a specific population, e.g. a study
targeting
young children with a rare disease.
This can cause major delays for the trial: in 2012, 80\% of trials in the US were  
aborted because of the lack of fitting participants \citep{Johnson2015AnEA}. 
While enrollment into the trial is usually proposed by a medical practitioner to an already known patient, new online recruitment solutions\footnote{See for instance \href{https://www.klineo.fr}{Klineo} or \href{https://digitalecmt.org/}{DigitalECMT}.} are promoting trials directly to patients who might not be familiar with clinical trials.
These solutions could help speed up and reduce the cost of the patient recruitment process, allowing to recruit hard-to-reach populations,
and target underrepresented populations
\citep{online_ctr}. 

In this study, we focus on patient recruitment for clinical trials by adopting the patient's point of view, thus using patient language (PL) to describe the patient's medical profile.
To enable the research community to explore this setting, 
we design a novel task, Natural Language Inference for Patient Recruitment (NLI4PR).
We create a dataset derived from patient profiles from the shared task TREC 2022 Clinical Trial Track (TREC-CT 2022) \citep{Roberts2022OverviewOT} and clinical trials' eligibility criteria for which the patient would be eligible or excluded. We frame the recruitment task into a Natural Language Inference (NLI) task. 
Our aim is to evaluate models' ability to infer from a given premise (the trial's eligibility criteria) whether the statement (the patient's medical profile) is entailed or contradicts the given premise. If there is an entailment, the patient can be enrolled in the trial; otherwise, the patient does not match the trial's eligibility criteria.
Since Large Language Models (LLMs) have demonstrated competitive results in similar shared tasks \citep{jullien-etal-2023-semeval, jullien-etal-2024-semeval},
we evaluate how they fare on the present new task.
Our contributions are the following:
\begin{itemize}
    \item Using patient language instead of medical doctor's language to describe the patient's medical profile and perform the patient-trial matching task.
    \item Creating a new dataset and task, NLI4PR, 
    aiming at matching patients to clinical trials using patient language. 
    \item Evaluating and comparing Large Language Models on the patient-matching task using medical and patient language.
\end{itemize}

\section{Related Work}
\label{sec:RW}

\subsection{Natural Language Processing for recruiting patients for clinical trials}

\begin{figure}
    \centering
    \includegraphics[width=\columnwidth]{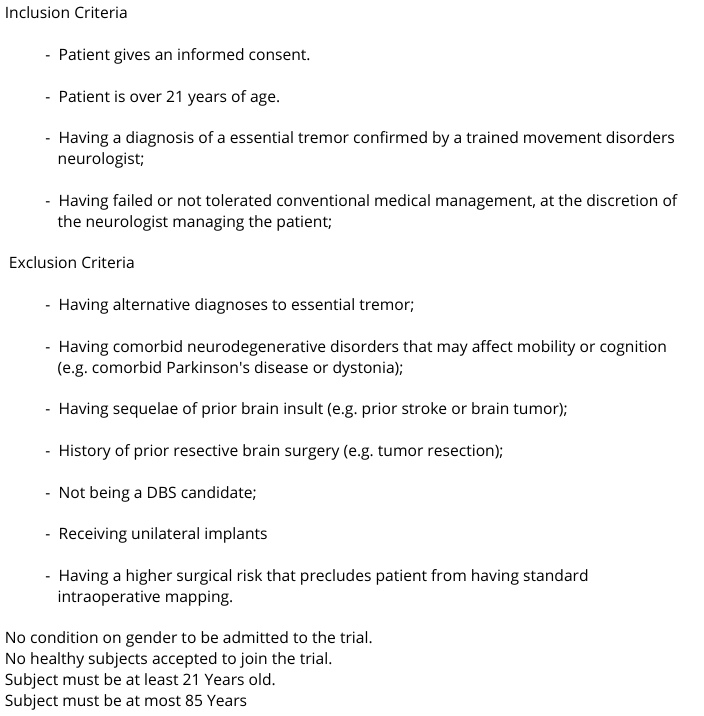} 
    \caption{Example of a CTR's eligibility criteria. Taken from NCT04581941, available on clinicaltrials.gov} 
    \label{fig:ex_criteria}
\end{figure}

Recruiting patients for clinical trials can be challenging and time-consuming. This is one of the main causes for trials to fail \citep{KANTOR2024104139}. Trials target a certain population, defined through the eligibility criteria designed at the beginning of the study (see Fig.~\ref{fig:ex_criteria}). These criteria are expressed as free text in the Clinical Trial Reports (CTRs). The traditional way of promoting trials to patients was made directly by healthcare professionals to known patients that might fit the trial. However, this involves a long manual review of patient profiles, which can also lead to screening errors. Thanks to the digitization of patients' medical records, called electronic health records (EHRs), systems based on NLP \citep{NLP_app1, NLP_app2} aimed at providing support to solve the patient-trial matching task.
These systems allow the automatic review of patients' profiles and trial eligibility criteria. They can either follow the trial-to-patients paradigm (for a given trial, the system suggests several patient profiles) or patient-to-trials (for a given patient, the system proposes several trials). 

The TREC-CT 2021 \citep{trec_2021}, 2022 \citep{Roberts2022OverviewOT}, and 2023 \citep{Soboroff_2024} shared tasks promote the development of NLP-based systems that address the patient-trial matching problem. These tasks provide patient topics, which are a short description of a patient's medical profile, in free-text form in the 2021 and 2022 editions or as structured text (as questionnaires) in the 2023 edition. The goal is to provide for each patient a ranked list of CTRs for which the patient would be eligible, excluded, or not relevant. With the recent advent of Large Language Models,
methods
using these models \citep{jin2024matching, trial_llama, Wornow2024ZeroShotCT} have been developed to perform the patient-trial matching. These methods have demonstrated competitive results compared to previous methods based on Masked Language Models. 

Natural Language Inference is a task that aims to determine whether a statement can be inferred from a given premise. This task is quite challenging as it requires different kinds of knowledge,
and involves finding evidence in the given pieces of text and confronting these pieces of evidence all together in order to conclude if there is an entailment or a contradiction.
The NLI4CT task \citep{jullien-etal-2023-nli4ct} uses NLI on clinical trials for various applications. Clinical trials are used as NLI premises, and statements have been manually generated. One of the targeted applications is patient recruitment, but the statements are using doctor's medical language. NLI4CT offers a benchmark to evaluate models on their common-sense, numerical, and biomedical abilities applied to the clinical trial domain. Besides, these premises not only consist of the eligibility criteria section, but also, in some instances, they consist of result,  intervention, or adverse events sections. Systems like that of  \citet{DeepEnroll} use NLI to model the patient recruitment task, using a fragment of the patient's EHR as the statement and the trial's eligibility as a premise. 
All of these approaches are based on the patient's EHR or other medical documents, and never on the patient's medical profile using patient language in a free-text form.
Our task is the first to propose an approach using patient language to match patients to clinical trials. 

\subsection{Processing Patient Language}  
\label{sec:POL}

According to \citet{seiffe-etal-2020-witchs}, a medical, technical term is either used by a physician or comes from Latin or Greek; a lay term is a term that can easily be understood by patients or is based on everyday language. Here, we define patient language (PL) as the expressions, terms, and formulations expressed in natural language that patients use to talk about their health and any health-related topic, which is broader than the definition proposed by \citet{seiffe-etal-2020-witchs}. 
Processing such language poses different challenges from those in traditional medical texts. While medical language uses precise terms to describe a concept, patients will use less precise expressions due to a lower level of medical knowledge, which often causes the patient's text to be inaccurate and also longer compared to one written by a healthcare professional. 
The patient's medical language is also highly influenced by their health literacy, often depending on their social background, age, and education level. PL also often conveys a load of negative emotions, such as fear, worry, anger, or anxiety \citep{Anderson2008WhatCM}.
In written text, typos and misspellings can also occur.   
Lay terms (or plain English) bridge the gap between the jargon of a complex domain and ``everyday life'' language. In the medical domain, they allow patients to make informed decisions, as for instance in the README dataset \citep{yao-etal-2024-readme}, which aims to provide patients with definitions for technical terms found in their EHRs in lay terms.
Medical to lay term glossaries have been created, such as that from the University of Michigan\footnote{\url{https://medicaldictionary.lib.umich.edu/}} or that of the European Medicines Agency.\footnote{\url{https://www.ema.europa.eu/en/documents/other/ema-medical-terms-simplifier_en.pdf}} MedlinePlus\footnote{\url{https://medlineplus.gov/}} \citep{medline_plus} also provides a glossary of medical concepts explained using lay terms and other synonyms. The Unified Medical Language System (UMLS) \citep{Bodenreider2004TheUM} is a set of health, biomedical-related vocabularies and standards for the medical domain. In particular the Consumer Health Vocabulary (CHV) provides some medical term to lay language mappings.

Usually, the goal behind the use of lay language is to summarize \cite{giannouris-etal-2024-plain} or simplify \citep{Attal_Ondov_Demner-Fushman_2023} the original technical text. \citet{giannouris-etal-2024-plain} summarized clinical trial reports with lay language to make them more easily accessible to non-experts but did not address the recruitment process.
In this paper, we do not try to summarize or simplify the patient profile but we use lay terms to study whether patient language is processed as well as medical technical language in clinical trial matching, so that patients themselves could be the starting point of recruitment for clinical trials.

\section{Corpus Creation}
\label{sec:corpus}
\begin{figure}[h]
    \centering
    \includegraphics[width=0.99\columnwidth]{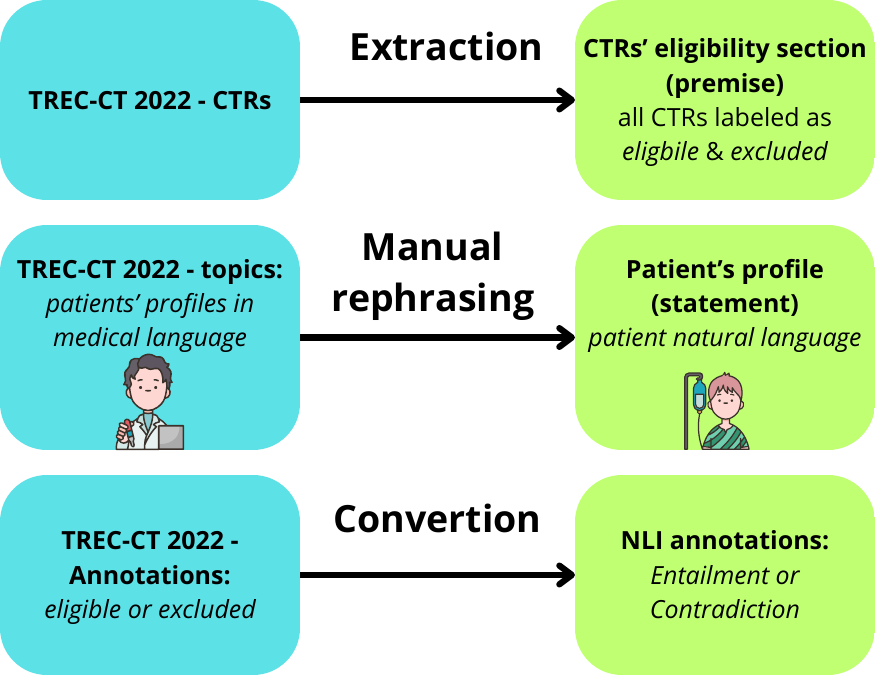}
    \caption{Corpus creation steps}
    \label{fig:creation_schema}
\end{figure}

To the best of our knowledge, no dataset exists in which lay language descriptions of patient profiles are used to identify matching clinical trials.
We therefore decided to create one.
To do so, we employ a 3-step process: (i) we start from TREC-CT 2022's patient topics, which express patient profiles in free-text, medical language. We then rephrase these topics using patient language (see Sec.~\ref{sec:rephrasing}). (ii) We collect the CTRs labeled as \emph{eligible} and \emph{excluded} in TREC-CT. Finally, (iii) we convert the task into a 2-way NLI classification task (see Sec.~\ref{sec:annotation}). Figure~\ref{fig:creation_schema} summarizes the process.

\begin{table*}
  \centering
  {\small\setlength{\tabcolsep}{4pt}
  \begin{tabular}{p{.3\linewidth}p{.3\linewidth}l}
    \hline
    \textbf{Medical term} & \textbf{PL example} & \textbf{Rephrasing strategy} \\
    \hline
    ALS (amyotrophic lateral sclerosis)     & sclerosis &   MedlinePlus + name simplification    \\
    ear discharge     &  fluid in my ear    &   MedlinePlus' description  \\
    hearing loss     & I could not hear as well as I used to    &  Paraphrase of the symptoms    \\
    His father died suddenly at age 35. & My dad died suddenly when he was 35, so I'm kind of scared. & Add emotion (fear) \\
    dyslipidemia     & cholesterol   &   MedlinePlus (Alternative names section)    \\
    Allopurinol     & Zyloric  &  MedlinePlus (Brand names section)       \\ 
    \hline
  \end{tabular}%
  }
  \caption{Examples of medical and patient language (PL) equivalents used in our task and the corresponding rephrasing strategy employed.}  
  \label{tab:POL_med_comparison}
\end{table*}

\subsection{Rephrasing into Patient Language}
\label{sec:rephrasing}

We used the 50 TREC-CT 2022 patient topics that describe the patient's last medical visit (emergency room, clinic, or primary care physician). Topics are written using medical language. Following MIMIC-IV's \citep{Johnson2023MIMICIVAF} descriptors, the patient topics contain the following information: chief complaint, history of present illness, patient demographics (age and gender), physical exams, and discharge diagnoses. Topics cover various diseases, such as genetic, endocrinal, or dermatological diseases, with patients presenting various profiles, from newborns to the elderly. 
To obtain PL topics, we tried two different approaches. The first consists of using Large Language models to rephrase the topics automatically. We tried with GPT-4o \citep{openai2024gpt4technicalreport} and Llama3-8B-Instruct \citep{grattafiori2024llama3herdmodels} and applied a simple prompt, displayed in Appendix~\ref{sec:rephrasing_tests_llm}. Both models seemed to grasp most of the information and adopt a patient perspective, using lay terms and the appropriate tone. However, they sometimes tended to remove quite important information (in the example displayed in Appendix~\ref{sec:rephrasing_tests_llm}, in both cases, gender is missing). To avoid these issues, we discarded this approach and opted for the approach below. 

\begin{figure*}
    \centering
    \includegraphics[width=.75\linewidth]{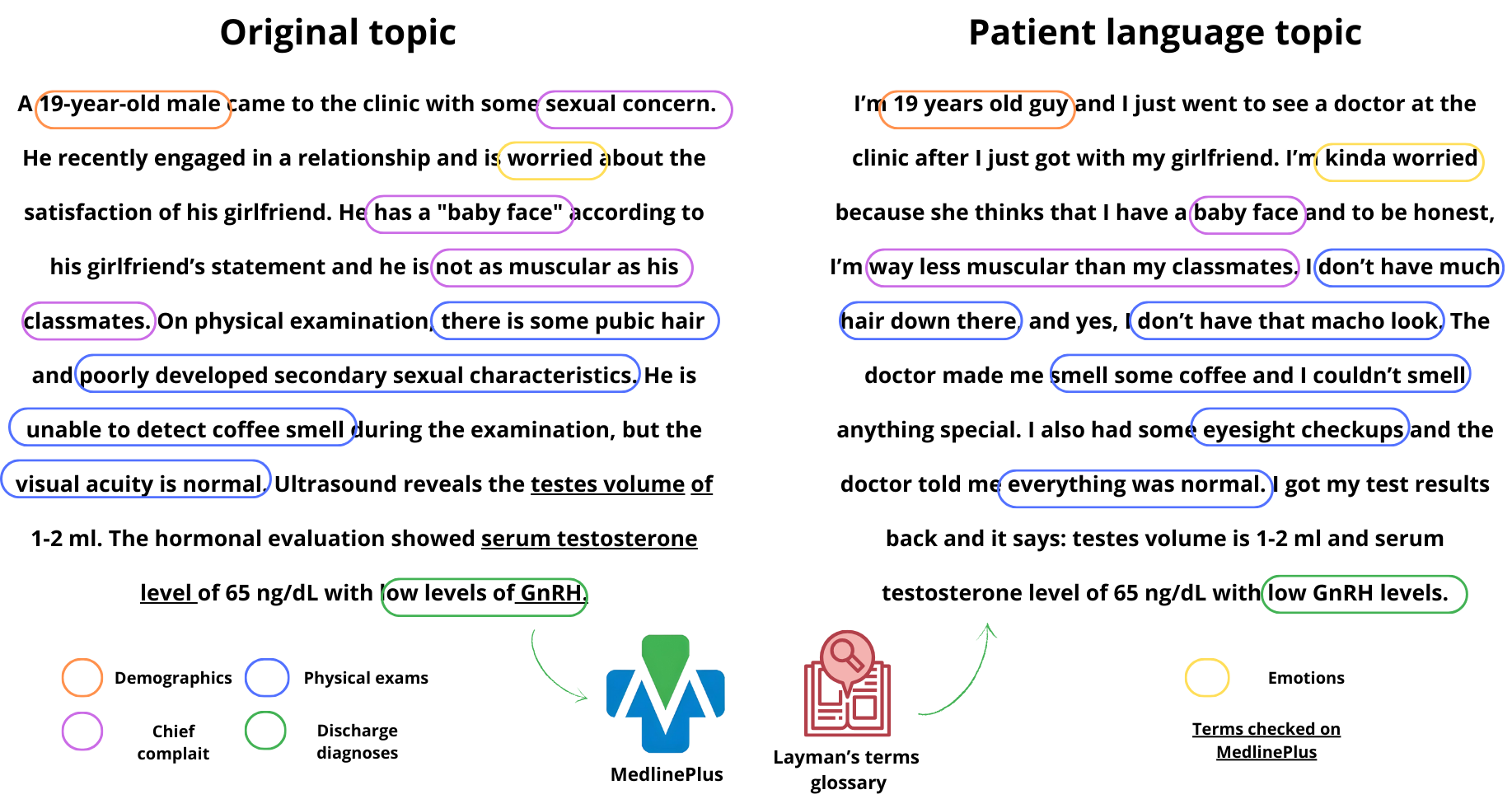} 
    \caption{Rephrasing of a patient topic, following MIMIC-IV categories and using MedlinePlus.}
    \label{fig:example_rephrasing}
\end{figure*}

To ensure consistency in the information contained in the topics, the first author manually rephrased the topics.
This author is experienced in working on medical texts and performing annotation tasks on medical documents, but does not hold any medical degree. We estimate that this level of expertise is suitable for our task since we are trying to represent the health literacy of an average patient. 
To get a better grasp of different patients' writing styles, we first conducted a manual evaluation with 6 human annotators presenting various patient profiles, described in detail in Appendix~\ref{sec:rephrasing_tests_patients}. We adopted a language similar to the one used by the participants. Apart from mapping the concepts from medical language to PL, we noticed that patients tend to use expressions representing their emotions, usually referring to fear, worry, or anxiety. We took this aspect into consideration in the rephrasing. Figure~\ref{fig:example_rephrasing} gives an example of the rephrasing process: 

\begin{enumerate}
    \item Selecting the important concepts in the original patient topic (following the MIMIC-IV categories mentioned before).
    \item Converting these concepts into patient language by using MedlinePlus for concepts unknown to the annotator or by using a lay-to-medical terms glossary. For each medical term, the annotator checks first MedlinePlus to understand the concept and look for lay language equivalents. They also check lay-to-medical glossaries to see other existing terms (although these glossaries often fall short for specific terms). If no equivalent was found in glossaries and MedlinePlus, the annotator paraphrases the term.  
    \item Styling the text using words that reflect the patient's emotions, by using adjectives that reflects fear, worry, etc. and by using exclamatory sentences. Additionally, we also tried to adjust language to the patient's age.
    \item Proofreading to ensure consistency with the original topic.
\end{enumerate}

To guide the rephrasing process, the annotator produced topics following this instruction (similar to the one given to the participants of the manual evaluation): "\emph{Describe the purpose of your last doctor appointment, the tests undergone, the obtained results or diagnosis as well as your age, gender, and past medical history. All in no more than a dozen sentences.}".

Table~\ref{tab:POL_med_comparison} displays a small sample of reformulations of the initial medical terms. 
To analyze a few linguistic features of the NLI4PR dataset, we compute readability and similarity metrics. Using some of the scores of BioLaySumm 2024 \citep{goldsack-etal-2024-overview}, we computed BERTScore \citep{BERTScore} for similarity between the patient and medical version of the topics, Flesch-Kincaid Grade Level (FKGL) \citep{fkgl}, Coleman-Liau Index (CLI) \citep{cli}, and Dale-Chall Readability Score (DCRS) \citep{dcrs} scores for readability.  
Tab.~\ref{tab:fact_read_metrics} reports the results of the different metrics.
\begin{table}
  \centering
  {\small
  \begin{tabular}{lcc}
    \hline
    \textbf{Metric} & \textbf{Patient} & \textbf{Medical}  \\
    \hline
    BERTScore & \multicolumn{2}{c}{89.5\%} \\
    \hline
    FKGL     & 6.24 & 8.83 \\
    DCRS    & 8.13 &  10.89 \\ 
    CLI     & 5.88 & 10.76 \\
    \hline
  \end{tabular}%
  }
  \caption{Similarity (BERTScore) between patient and medical versions of the topics. Readability (FKGL, CLI, and DCRS) measures for patients vs medical versions of the topics.}
  \label{tab:fact_read_metrics}
\end{table}

The patient and medical topics still keep similar features with a high BERTScore of 89.5\%.
For the patient language topics, FKGL and DCRS scores both respectively indicate that a 11-17 years old student and a $11-12^{th}$ grade student could understand the topics written in patient language. Although, the CLI measure estimates the readability to be accessible for a 5-$6^{th}$ grader. However the topics produced are accessible to the majority of the population and correspond to what we would expect from an adult's average health literacy. For the medical version of the topics, the scores are higher (2.5 points more for FKGL and DCRS) and almost doubled for CLI, bringing the readability level to a 10-$11^{th}$ grade student.  
To see if the proportion of medical terms is more important in the medical version of the topics, 
we used QuickUMLS \citep{Soldaini2016QuickUMLSAF} to extract medical concepts indexed in UMLS. 
For 92\% of the topics, the medical language version contains more terms taken from the UMLS than its patient language equivalent. On average, patient language topics contain 21 terms taken from the UMLS versus 25 for the topic's medical version. Although the average length of a patient language topic is 116 words versus 98 for medical language. This suggests that patient indeed tend to use paraphrase to refer to medical terms. 

\subsection{Conversion into an NLI task}
\label{sec:annotation}

TREC-CT's original aim is to rank a large number of CTRs in terms of eligibility for a given patient topic. 
There are 3 ranking levels: \emph{eligible} (the patient described in the topic can take part in the trial), \emph{not relevant} (the trial's eligibility criteria do not seem relevant for the patient described in the topic and there is not enough information to qualify for the trial), and \emph{excluded} (the patient described in the topic does not match the trial's eligibility criteria). 
Natural Language Inference aims to determine whether a statement entails a given premise, thus in our context, whether the patient topic (statement) entails the trial's eligibility criteria (premise). 
We map TREC-CT's annotations to NLI annotations: \emph{eligible} is mapped to \emph{entailment}, and \emph{excluded} to \emph{contradiction}. We did not map the instances labeled as \emph{not relevant} to \emph{neutral} as TREC-CT's goal was to rank trials by relevance and not to test patients' eligibility. 
We describe the internal inference process that should be employed in order to predict the right label. The patient topic $Pat$ has a set of $n$ features $f$ (age, disease, gender, etc.): $ Pat = \{f_1, ..., f_n\} $. The eligibility section is composed of $m$ inclusion criteria $Inc$ and $k$ exclusion criteria $Exc$: $Inc = \{i_1,...,i_m\}$ and $Exc=\{e_1,...,e_k\}$. We define the inference relationship between the statement $Pat$ and the premise $Inc, Exc$ as:

\begin{equation}
   \forall i \in Inc, \exists f \in Pat; entail(i, f)
\end{equation}

\begin{equation}
    \forall e \in Exc, \forall f \in Pat, contradict(e, f)
\end{equation}

\begin{equation}
    (1) \wedge (2) \Rightarrow Entailment 
\end{equation}
where \emph{contradiction} holds if \emph{entailment} does not. In other words, the model has to infer that for every feature $f$ of a patient, it entails with every inclusion criteria and that it contradicts with every exclusion criteria, for the model to output \emph{Entailment} as the final prediction. 

For each topic, we extract all the CTRs labeled as \emph{excluded} and \emph{eligible} in TREC-CT, resulting in, for each patient topic, several \emph{(patient topic, CTR)} pairs labeled either with \emph{entailment} or \emph{contradiction}. Our resulting task is a 2-way NLI classification task.

\subsection{Resulting dataset}

The resulting dataset consists of 7007 instances, split into training, development, and test sets (representing 70\%, 10\%, and 20\% of the whole dataset, respectively).
3939 are labeled as \emph{Entailment} and 3068 as \emph{Contradiction}. Table \ref{tab:dataset_metrics} displays the number of instances per split and the label distribution. We provide two kinds of statements: \emph{statement\_medical}, which is the original TREC-CT's patient topic (in medical language), and \emph{statement\_pl}, which is the PL rephrased topic. The \emph{premise} field is composed of the extracted eligibility section of the CTR. Additionally, we provide the study's title in the \emph{NCT\_title} field and its corresponding id in \emph{NCT\_id}. 
As in \citet{jullien-etal-2023-nli4ct}, our dataset involves several challenges: biomedical reasoning, numerical reasoning, and commonsense reasoning.  
Appendix \ref{sec:corpus_stats} displays more statistics. 
The dataset is freely available on HuggingFace.\footnote{\url{https://huggingface.co/datasets/Mathilde/NLI4PR}}

\begin{table}[ht]
    \centering
    \begin{tabular}{lcc}
        \hline
        \textbf{Split} & \textbf{\# Entailment} & \textbf{\# Contradiction} \\
        \hline
        Train     &    2757    &   2147 \\
        Dev     &  295    &  230 \\
        Test   &   887 & 691    \\
        \hline
    \end{tabular}
    \caption{Distribution of \emph{Entailment} and \emph{Contradiction} instances in the dataset splits.}
    \label{tab:dataset_metrics}
\end{table}

\section{Methods}
\label{sec:methods}
Using this new dataset, we perform initial experiments to evaluate the ability of LLMs to solve the task with lay- vs.\ medical-language patient profiles.

\begin{figure*}
    \centering
    \begin{subfigure}[b]{\columnwidth}
        \includegraphics[width=0.8\columnwidth]{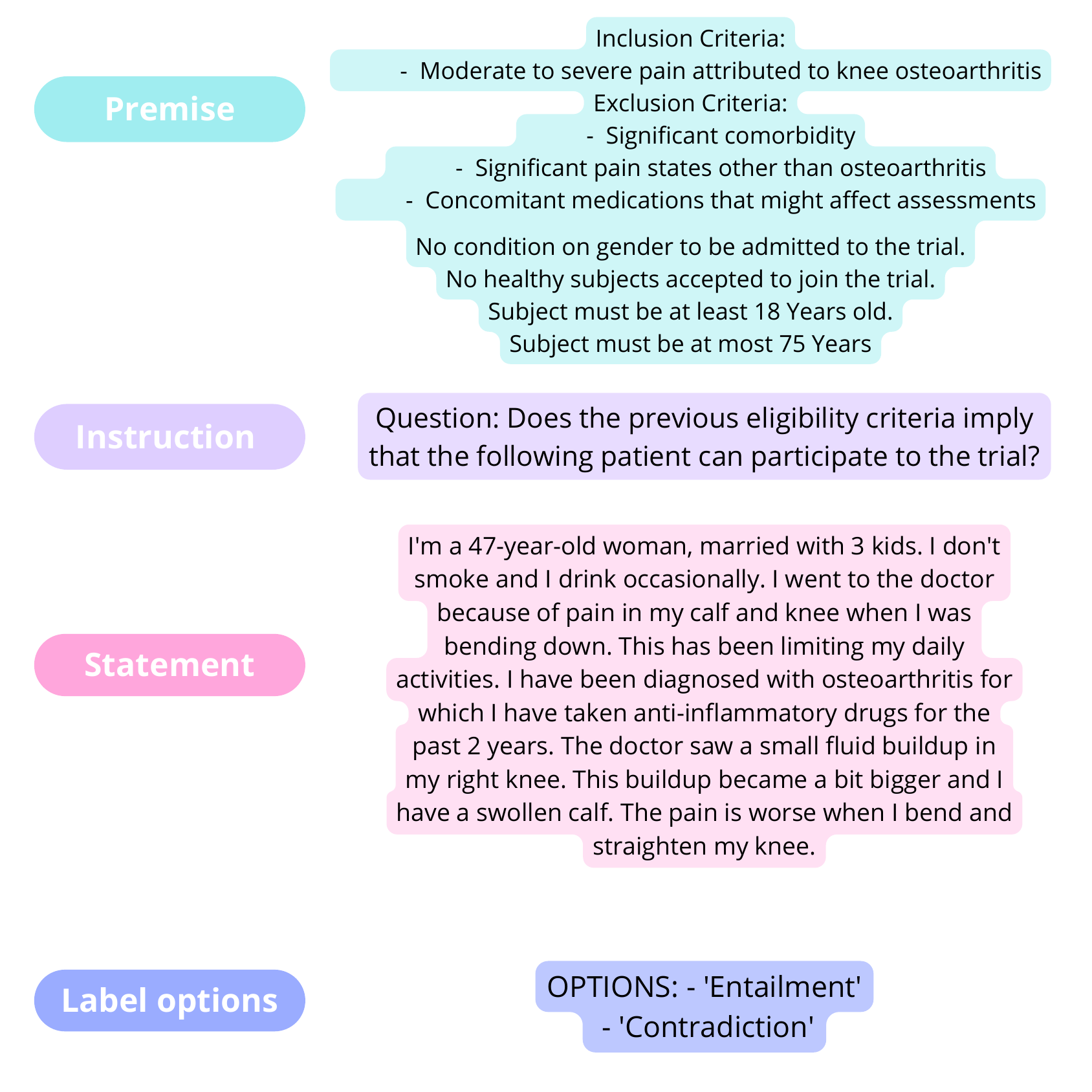}
        \caption{Example of a prompt using the \emph{Vanilla} template}
        \label{fig:prompt_vanilla}
    \end{subfigure}
    \hfill
    \centering
    \begin{subfigure}[b]{\columnwidth}
        \includegraphics[width=0.8\columnwidth]{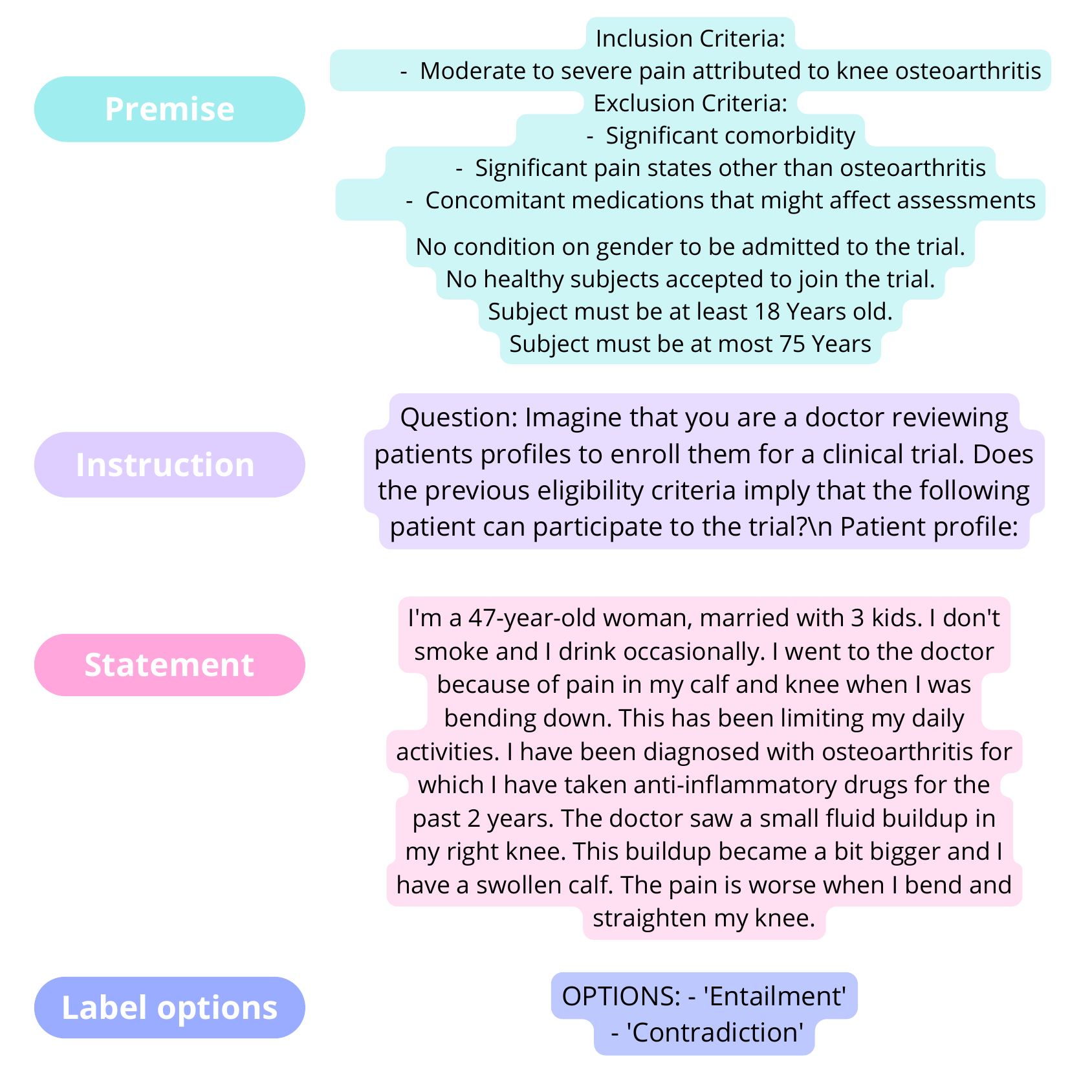}
        \caption{Example of a prompt using the \emph{Persona}-style template}
        \label{fig:prompt_persona}
    \end{subfigure}
\end{figure*}

We prompt four open-source Large Language Models 
using two prompting templates. The first template, \emph{vanilla}, is made of a simple instruction described in Figure \ref{fig:prompt_vanilla}; the second template, \emph{persona}, aims at impersonating the model into a medical practitioner reviewing patient profiles and deciding whether they can participate into the trial or not (see Figure \ref{fig:prompt_persona}).

The templates are structured as follows: the \emph{premise}, which is the eligibility criteria section of the clinical trial, the instruction, the \emph{statement}, which is the patient profile, either expressed in PL or using medical language, and finally we provide the possible answers, \emph{Entailment} or \emph{Contradiction}. We perform all the experiments in a zero-shot setting, meaning that we do not show any previous demonstration to the model.  

We use models that previously achieved competitive results on the similar SemEval task of NLI4CT:
\begin{itemize}
    \item Flan-T5-XXL \citep{flan}, an 11 billion parameters instruction-tuned sequence-to-sequence model.
    \item Qwen2.5-7B-Instruct and Qwen2.5-14B-Instruct \citep{qwen2.5}, instruction-tuned decoder-only models respectively with 7 and 14 billion parameters.
    \item Mixtral-8x7B-Instruct-v0.1 \citep{Jiang2024MixtralOE}, a 45 billion parameters decoder-only model pretrained using a mixture of experts approach.
\end{itemize}
These models are all pretrained on general domain data. As in \citet{jullien-etal-2023-nli4ct}, we choose macro-F1 score as the evaluation metric. We perform the evaluation on the whole test set. We use a temperature of 0.7, a top\_p of 1 and top\_k of 0. For comparison, we compute the majority baseline corresponding to the case where all the predictions would be labeled as \emph{Entailment}. Comparison is also done against a random classifier where the seed used is 42.

\section{Results}
\label{sec:results}

Table \ref{tab:results} displays the results obtained by the models on the two types of templates.

\begin{table}
  \centering
  {\small
  \begin{tabular}{lcccc}
    \hline
    \textbf{Model} & \textbf{Lay-\emph{V}}  & \textbf{Lay-\emph{P}} & \textbf{Med-\emph{V}}  & \textbf{Med-\emph{P}} \\
    \hline
    Majority & \multicolumn{4}{c}{36.0} \\
    Random & \multicolumn{4}{c}{50.0} \\  
    \hline
    Flan-T5-XXL     & \underline{66.0} & 61.8 &   \underline{72.1}   &   {67.5}   \\
    Qwen-7B     & 64.1 & \underline{62.9} &   65.5    &  64.7  \\
    Qwen-14B     & \textbf{71.8} & \textbf{69.8} &  \textbf{73.1}     &   \textbf{73.7}  \\
    Mixtral-8x7B   & 60.7 & 56.5 &   70.8  &    \underline{71.2}  \\
    \hline
  \end{tabular}%
  }
  \caption{Macro F1 score (in \%) for the different baselines, using our different prompting templates in a zero-shot setting, on the test set.
  \emph{Lay} is \emph{patient} language,
  \emph{Med} is \emph{medical doctor}'s language,
  \emph{V} stands for \emph{vanilla} template and
  \emph{P} stands for \emph{persona} template.
  The majority baseline is \emph{Entailment}. Seed for the random baseline is 42.}
  \label{tab:results}
\end{table}

Qwen-14B achieves the best results for all kinds of templates, up to 37.7 points higher than the majority baseline and 23.7 for the random baseline. All models perform better on medical language than on PL. We believe this loss of performance may come in part from the lack of precision of layman terms used in PL, in comparison to medical terms that define a more precise concept. When trying to match eligibility criteria, the model might not be able to determine the patient's eligibility if in the PL statement, the concept is not precise enough. E.g., in the following example, the eligibility criteria states \emph{"Subjects having a diagnosis of probable or definite \textbf{ALS} in accordance with the Revisited El-Escorial Criteria."}, the patient topic in medical language uses the acronym \emph{ALS}, however in the patient topic in PL, the term used is simply \emph{sclerosis} (see Table \ref{tab:POL_med_comparison}). With PL, the model cannot determine which type of sclerosis the patient is suffering from and thus might not match it to the trial. 

Using a persona template did not necessarily lead to better results; Flan-T5 performed even worse when using PL.  
Despite being the larger model, Mixtral is the worst-performing when using PL, and in the worst case being only 6.5 points above the random baseline. In the case of Qwen, more parameters (increasing from 7B to 14B) improved performance, with a gain of up to 9 points for the Med-P template.

\section{Error Analysis}
\label{sec:error_analysis}

\paragraph{Medical vs Patient Language}
\label{sec:med_vs_pat}

We examined on which patient topic models tend to fail, either using PL or medical language: for this purpose, we compute the misclassification rate (MCR) for each patient topic $t$ using the predictions of each model and the gold standard: 

\[MCR(t) = \frac{misclassification\_topic\_t}{total\_count\_topic\_t}\] 
We compute MCR for all topics with all models' predictions across all templates, where $misclassification\_topic\_t$ is the number of misclassifications for topic $t$ and $total\_count\_topic\_t$ the number of instances using topic $t$ as the statement in the dataset.  
We derive $MCR_{pl>med}$ where the models perform better with topics using medical language than PL, and conversely $MCR_{med>pl}$ where models were better using patient language, for each patient topic $n$:

$$ MCR_{pl-med}(t) = MCR_{pl}(t) - MCR_{med}(t) $$

$$ MCR_{pl>med} = \max_{t \in [1,50]} MCR_{pl-med}(t)  $$

$$ MCR_{med>pl} = - \min_{t \in [1,50]} MCR_{med-pl}(t) $$

Across all the models, the patient topic occurring the more often for $MCR_{pl>med}$ is patient \#21, and the one for $MCR_{med>pl}$ is patient \#30. Appendix~\ref{sec:patients_21_30} displays both patients' profiles. The descriptions of patient \#21 in medical language and PL are similar in terms of demographics, chief complaint, and physical exams. However, for the discharge diagnosis, medical language mentions \emph{ALS} while PL mentions \emph{sclerosis} (see Tab.~\ref{tab:POL_med_comparison}), thus not mentioning the specific kind of sclerosis diagnosed. 
For patient \#30, the physical exam observations have been greatly simplified in the PL version. Otherwise, information remains consistent with the medical language version. 

We quickly investigate if these differences can be the reason for misclassification. We allow Qwen-14B to output a longer sequence of tokens when prompted with a single example of patient \#21 and \#30 (see Appendix~\ref{sec:template_justifs}). Qwen provides a brief explanation of the reason for its prediction. We compare the justifications given for the \emph{Lay-V} and \emph{Med-V} prompts. 

For \#21, Qwen predicted the right label (\emph{Entailment}) for \emph{Med-V} and the wrong label for \emph{Lay-V}. The misclassification comes from the case depicted in Sec.~\ref{sec:results}. Qwen mentioned that \emph{sclerosis} does not necessarily involve an ALS, which is technically true. PL lacks precision, which can lead to misclassification, whereas the model can predict the right label with medical language for the same case.  

For \#30, Qwen predicted the right label (\emph{Entailment}) for \emph{Lay-V} and the wrong label for \emph{Med-V}. The patient topic describes a woman suffering from osteoarthritis. In order to solve the inference, the model has to perform numerical inference to determine if her age fits the age range of the inclusion criteria, check that the diagnosis of osteoarthritis fits with the inclusion and exclusion criteria and that the patient does not suffer from other disorders. For \emph{Lay-V} the model reports that it compared the age range, the osteoarthritis diagnostic with the eligibility criteria. For \emph{Med-V} the model got misled by one of the symptoms and inferred another potential disease, that could fall under one exclusion criterion. In this case, having more information that was not directly linked to the criteria confused the model and led to a wrong prediction.

\begin{figure}
     \centering
     \begin{subfigure}[b]{\columnwidth}
         \centering
         \includegraphics[width=\columnwidth]{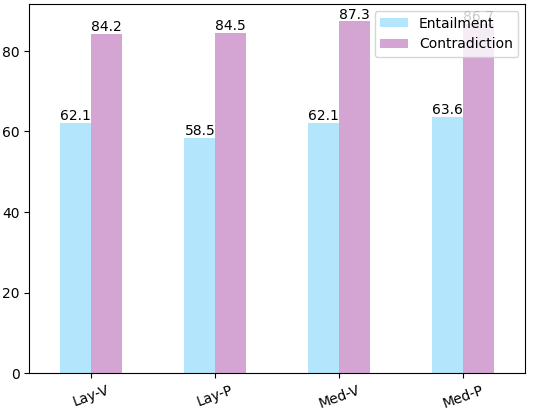}
         \caption{Entailment and Contradiction accuracy for Qwen-14B's predictions.}
         \label{fig:ent_vs_cont_qwen}
     \end{subfigure}
     \hfill
     \begin{subfigure}[b]{\columnwidth}
         \centering
         \includegraphics[width=\columnwidth]{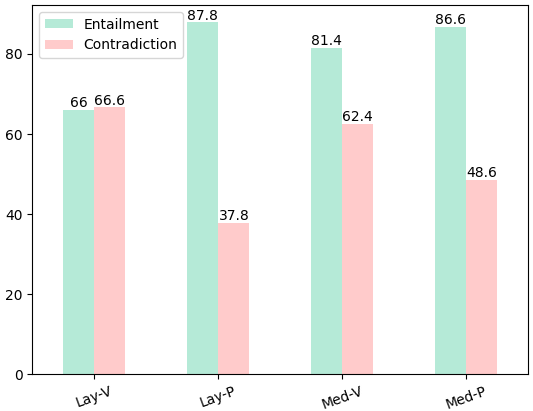}
         \caption{Entailment and Contradiction accuracy for Flan-T5-XXL's predictions.}
         \label{fig:ent_vs_cont_flan}
     \end{subfigure}
        \caption{\emph{Lay} is \emph{patient} language,
          \emph{Med} is \emph{medical} doctor's language,
          \emph{V} stands for \emph{vanilla} prompt and
          \emph{P} stands for \emph{persona} prompt.}
        \label{fig:ent_vs_cont}
\end{figure}

\paragraph{Which is harder, \emph{Entailment} or \emph{Contradiction}?}

We compute the accuracy per label for the two best-performing models, Qwen-14B and Flan-T5-XXL (see Figure \ref{fig:ent_vs_cont}). Qwen is performing up to 26 points better on \emph{Contradiction} than on \emph{Entailment}. This behavior is consistent with all the types of templates. Surprisingly, Flan-T5 obtains up to 50 points more in predicting \emph{Entailment} than \emph{Contradiction}, and this observation applies to all templates except \emph{Lay-V}.
Predicting \emph{Contradiction} seems rather simple compared to predicting \emph{Entailment}. Since a patient would not be eligible for a clinical trial if their characteristics do not comply with at least one of the exclusion criteria, this would lead to a direct assertion of the \emph{Contradiction} label. Meanwhile, for \emph{Entailment}, the model has to go through all the patient's features and compare them to all the inclusion and exclusion criteria, which involves more knowledge and computations.

\section{Future Work and potential applications}

One direction for future work would be to fine-tune the models (using our training and development sets) to see if it would improve performance. 
Systematically evaluating models' explanations would also allow to determine if the model is predicting the right label for the right reason and, hence, detecting the right pieces of evidence in the text to make its prediction. This evaluation could be done using the LLM-as-a-judge paradigm \citep{llm-as-judge}, where one or several LLMs could evaluate if the retrieved evidence and explanations are correct.    
Expanding the dataset with new patient profiles with various health literacy levels and diseases would also allow to evaluate the models on more diverse cases. 

We hope this work can pave the way to the development of more NLP applications to promote clinical trials directly to patients, using their own language. We believe that proposing these kinds of interfaces would allow to reduce the recruitment workload and to promote trials to a wider population.

\section{Conclusion}
In this study, we present a novel task, Natural Language Inference for Patient Recruitment (NLI4PR), that aims to use patient language to match patients to clinical trials. The patient-to-trial matching is usually done using a description of the patient in doctor's medical language. Here, we adopt another approach where the patient describes their own profile using their own language. Patient language presents major differences compared to doctor's medical language due to the patient's limited health literacy. We evaluated the ability of several open Large Language Models to deal with patient language and compare it to the use of medical language. We frame the task as a Natural Language Inference task.
For this, we create a new dataset derived from the patient profiles provided by TREC-CT 2022, and the clinical trials ranked as \emph{eligible} and \emph{excluded} in TREC-CT 2022. 

We found that all models obtained an F1 score much higher than the majority baseline on our test set, using medical language but also using patient language. Models struggled more with patient language than with medical language, however the gap between the two settings was rather low. We found that this gap in performance is mainly coming from the loss of precision in the terms used by patients compared to medical professionals.
We observe that most of the errors are coming from issues in enforcing the inclusion and exclusion criteria rather than the difference between the patient language and the doctors' medical language.

\section{Limitations}
The patient topics have been built to keep important patient information. We hypothesize that in a real-world scenario, a patient describing their medical history and condition might miss some elements, making the task even more challenging. In addition, despite the effort made to diversify the phrasings, the rephrasing has been done by a single annotator, which might limit diversity. An improved approach would be to ask a diverse pool of patients to produce their own statements to represent differences in pathology, social background, levels of health literacy, and phrasing. 

Since we perform a 2-way classification and we did not consider the cases labeled as \emph{not relevant} from TREC-CT, we cannot directly compare our results with those of TREC-CT participants.

\section{Ethical Considerations}

The patient profiles are taken from the original TREC-CT 2022 shared task. They do not contain any element or piece of information that could lead to identification of any individual. The rephrased version using layman's terms does not contain any personal information either. Clinical trials are extracted and processed from \href{https://clinicaltrials.gov/}{clinicaltrials.gov}. This resource is freely available, provided by the National Library of Medicine, and is an official U.S. Department of Health and Human Services website.

\section*{Acknowledgments}
 This work received funding from the CNRS through grant 80|PRIME and the French ``Agence Nationale pour la Recherche'' under grant agreement ANR-22-CPJ1-0087-01. This work was granted access to the HPC resources of IDRIS under the allocation 2024-AD011015277 made by GENCI.

\bibliography{custom}

\appendix

\section{Rephrasing patient topics using Large Language Models}
\label{sec:rephrasing_tests_llm}

Table \ref{tab:rephrasing} displays an example of the different outputs in PL obtained. The prompt used with the Large Language Models is the following:

\begin{quote}
    Rephrase the following patient profile like it would be expressed by the patient themselves: 
\emph{original\_topic}
\end{quote}

\begin{table*}
  \centering
  \begin{tabular}{c|p{12cm}}
    \hline
    \textbf{Source}           & \textbf{Output text}  \\
    \hline
    Original topic      &  A 19-year-old male came to clinic with some sexual concern. He recently engaged in a relationship and is worried about the satisfaction of his girlfriend. He has a "baby face" according to his girlfriend's statement and he is not as muscular as his classmates. On physical examination, there is some pubic hair and poorly developed secondary sexual characteristics. He is unable to detect coffee smell during the examination, but the visual acuity is normal. Ultrasound reveals the testes volume of 1-2 ml. The hormonal evaluation showed serum testosterone level of 65 ng/dL with low levels of GnRH. \\
    \hline
    ChatGPT     & I'm 19, and I recently started a relationship. I've been feeling a bit worried about whether I'm making my girlfriend happy. She mentioned that I have a 'baby face,' and I’ve noticed I’m not as muscular as most of my classmates. I also can’t smell coffee, which I realized during this visit. I do have some pubic hair, but I think I might be behind in developing other features that guys my age have. The doctor checked me out, and they said my testosterone levels are low, and my testicles are smaller than normal. They also mentioned something about low levels of another hormone, GnRH. \\
    \hline
    Llama-3-8B-Instruct     &  I'm 19 years old and I've been dating someone for a little while now. To be honest, I'm really worried that I'm not good enough for her. She says I have a 'baby face' and I'm not nearly as muscular as my friends. I feel like I don't measure up. When we're together, she sometimes mentions how unsatisfied she is with our intimate moments, which makes me even more anxious.
    Physically, I notice that I don't have much body hair or muscle development compared to others my age. During our last exam, I couldn't even smell the coffee because of my poor sense of smell! But at least my eyesight seems okay.
    I also just had an ultrasound done on my testicles, and the doctor told me they're smaller than usual. And when they checked my hormone levels, it turned out that my testosterone levels are super low - way lower than normal. It's making me feel really self-conscious and unsure about who I am and what I can do to make things better between me and my girlfriend. \\
    \hline
    Human &  I'm 19 years old guy and I just went to see a doctor at the clinic after I just got with my girlfriend. I'm kinda worried because she thinks that I have a baby face and to be honest, I'm way less muscular than my classmates. I don't have much hair down there, and yes, I don't have that macho look. The doctor made me smell some coffee and I couldn't smell anything special. I also had some eyesight checkups and the doctor told me everything was normal. I got my test results back and it says: testes volume is 1-2 ml and serum testosterone level of 65 ng/dL with low GnRH levels. \\
    \hline
  \end{tabular}
  \caption{\label{tab:rephrasing}
    Outputs obtained after rephrasing the original topic into PL. Using either LLMs or a human annotator.
  }
\end{table*}

\section{Rephrasing patient topics using patients}
\label{sec:rephrasing_tests_patients}

\paragraph{Goal} To have examples of the way patients express themselves, we asked 6 participants to produce statements using patient language. These participants are volunteers who are not authors of this paper. They all have different profiles in terms of health literacy, cultural backgrounds, educational backgrounds, and age. All the materials (persona, generated medical reports and guidelines) are available here\footnote{\url{https://github.com/CTInfer/NLI4PR/tree/main/human_evaluation}}.

\paragraph{Participants} We note $A_i$ a participant.

A1: 25-year-old Cambodian man with no medical training. He holds a Master's degree in computer science. His mother tongue is Khmer, but has a B2 English level. 

A2: 69 years old French man. He has a PhD in biochemistry but no particular training in medicine. His native language is French.

A3: 60-year-old French woman. She is a secondary teacher with no special training in medicine. Her native language is French. 

A4: 29 years old Greek woman. She is a PhD student majoring in NLP. She has no particular medical training. Her native language is Greek, but she has a C2 English level.

A5: 26 years old Chinese man. He is a PhD student in biology working on oncology. His native language is Chinese, but he has a B2 level in French. 

A6: 25 years old French woman. She is currently pursuing her 8th year of medical training, specializing in anesthesiology and intensive care. Her native language is French.

\paragraph{Settings} We designed our experiment in order to obtain as spontaneous as possible answers from the participants. We gave all participants a persona directly adapted from one of the TRECT-CT's patient topics. To avoid influencing the participants, the persona was built as a set of fields (age, gender, tests undergone, chief complaint, medical history, current medication) with keywords only. We also used ChatGPT to generate mock-ups of test results\footnote{For instance, a generated karyotype report  \href{https://github.com/CTInfer/NLI4PR/blob/main/human_evaluation/topic\%2013/Karyotype_Evaluation_Report_topic_13.pdf}{here}.} with the results mentioned in the original TREC-CT 2022 topic. We gave the following instruction to the participants: 

\begin{quote}
    Following the provided persona and test report, describe the purpose of your last doctor appointment, the tests undergone, the obtained results or diagnostic as well as your age, gender and past medical history. All in no more than a dozen of sentences.
\end{quote}

\paragraph{Results}
\textbf{A1:} ``I'm a 25 year old man. My roommate always says that mostly when we are watching TV in the afternoon, he noticed that I usually fall asleep even while we are talking. it's a bit embarrassing. I think I am not a person like that. I feel something is wrong. So I decided to meet a doctor to check about this. So the doctor asked me to do the sleep test which is called Multiple Sleep Latency Test. This test just requires me to take a bunch of short naps during the daytime and the doctor monitored my sleep. It is so surprising that the test also confirms that I quickly fall asleep and enter REM sleep. The doctor said this could be narcolepsy, which is a sleep disorder that causes excessive daytime sleepiness. I'm waiting to do an overnight sleep study and see a sleep specialist to figure out what the next steps are.''

\textbf{A2 (translated):} ``My medical check-up is not very encouraging. The blood test results are alarming. Serum antibodies are abnormally high, indicating an inflammatory state. The muscle biopsy confirms the presence of inflammation. The interpretations provided by the medical analysis laboratory suggest systemic inflammatory muscle damage. The doctor suspects dermatomyositis. However, he suggests further tests. In my quest to understand dermatomyositis, I learn that it is associated with various cancers. I think my life is about to be turned upside down if the doctor's diagnosis proves correct. Maybe I'll have to undergo chemotherapy and so on. At my age, my professional and emotional life could be seriously devalued. I'll decide after the next medical examinations.''

\textbf{A3 (translated):} ``I am a 33-year-old woman. I consulted a doctor because I'm persistently tired, have less appetite than before and have lost 4 kilos in two months.
At the hospital, tests showed generalized hyperpigmentation. A blood test revealed abnormally low levels of cortisol, a sign of “primary adrenal insufficiency”, a dysfunction of the glands above the kidneys that produce hormones. Could my health problem be due to my past history? I suffer from Hashimoto's disease, an autoimmune disorder caused by a thyroid disorder. I am treated for this disease by taking levothyroxine on a regular basis. But obviously, this medication no longer seems sufficiently effective. To resolve the adrenal insufficiency I'm suffering from, and thus stimulate the function of the adrenal glands, the doctors have prescribed a glucocorticoid-based treatment, on a long-term basis but under supervision. I will need to consult my GP regularly to observe the clinical symptoms - the extent of the brown spots on my skin - and to prescribe a blood test to monitor cortisol levels and, if necessary, adjust and rebalance the treatment. I was also recommended a parallel follow-up with an endocrinologist.''

\textbf{A4:} ``I am a 67 year old woman. I went to the ophthalmologist because I was not seeing well from both of my eyes especially during the night. The physical exam showed that my pupils are normal in diameter both in the light and darkness, however the acuity test results confirmed that I have a blurry vision of 50/100 that is probably linked to cataract. I will need to consult a second ophthalmologist to confirm the diagnosis and I might need to do further lab tests.''

\textbf{A5 (translated):} ``My name is Jean Martin and I'm 52 years old. On 22/01 I had two radiology examinations: a thoracic X-ray and an oeso-gastro-duodenal transit. The purpose of these examinations was to find an explanation for my symptoms of thoracic burning and acid reflux, which have been treated piecemeal with PPIs (proton pump inhibitors = anti-acids). I have no other antecedents than my obesity, I don't smoke or drink. Dr. Dupuis, a radiologist, interpreted these examinations and concluded that I had a hiatal hernia due to stomach sliding, with no signs of complications: no ulcerations, no digestive perforation and permeability of the lower esophageal sphincter, with no visualized esophageal reflux.
Treatment with ipp is indicated, as is follow-up by a specialist in gastroenterology. If the symptoms become too incapacitating, I'm advised to undergo 2nd-line laparoscopic surgery to reconstruct the stomach, which is still a major operation. I prefer to try medical treatment in 1st intention as agreed. I have been informed of the serious signs of my illness, which require me to undergo urgent appointment.''

\textbf{A6 (translated):} ``Hello, I'm currently 26 years old. I went to the clinic today because I felt down at the gym. I exercise often but it's been the 4th time that this happens. From time to time, I experience vertigo while I'm resting and I don't understand why. I exercise everyday and I don't have any other diseases for now. At the emergency room, the doctor asked me to do an X-ray and he showed me that I have a heart malformation. He told me that the volume of my left and right side are not equivalent. What's wrong? Should I stop working out?''

\paragraph{Conclusion}  Most participants followed the instructions correctly or at least partially (A2 forgot to mention their age and gender). 
A1, A2, A3, and A5 expressed some kind of worry regarding their symptoms and diagnosis, especially for A1 and A2, where the participants inquired about the consequences of their disease. All participants use reported speech to talk about their test results or the doctor's diagnosis. A3 and A4 directly cite some results directly taken from their test results. 
We observe that A2 and A3 did some supplementary research regarding their diagnosis (probably by searching their diagnosis in a search engine).

\clearpage
\section{Corpus statistics}
\label{sec:corpus_stats}

\begin{table}[ht]
    \centering
    \begin{tabular}{lc}
        \hline
        \textbf{Metric} & \textbf{Value}  \\
        \hline
        \# of CTRs (whole dataset)     &    6649        \\
        \# of CTRs (train dataset)    &  4713      \\
        \#  of CTRs (dev dataset)  &  523     \\
        \#  of CTRs (test dataset)  &  1564     \\
        \hline
    \end{tabular}
    \caption{Dataset metrics}
    \label{tab:extended_dataset_metrics}
\end{table}

\section{Patients \#21 and \#30}
\label{sec:patients_21_30}

\subsection{Patient \#21}

\noindent\textbf{Medical language:}
A 47-year-old man comes to the clinic for the follow up of his neuromuscular disease. He experienced gradual, progressive weakness of the left upper extremity over the last year.  Over the last few months, he has also noticed weakness in the right upper extremity.  BP is 120/75, PR is 80 and temperature is 37 C. Reflexes are brisk in the upper extremities, and the plantar responses are extensor.  Mild gait ataxia is present. The patient is under treatment of Riluzole 50 mg BID with the diagnosis of ALS.

\noindent\textbf{Patient language:}
I've been suffering from a neuromuscular disease for a while now, and I went to my doctor's office. I'm now a 47-year-old man and over the past year I experienced a progressive and gradual weakness of my left upper extremity, and over the past month, I also noticed a weakness over my right upper extremity. My heart rate was 120/75, and my  PR was 80 with 37°C for temperature. My reflexes are not good in my upper extremities, and I have trouble with my balance. I'm also under Exservan 50 mg for my sclerosis.

\subsection{Patient \#30}
\noindent\textbf{Medical language:}
A 47-year-old woman comes to the office complaining of pain in the calf and knee when she bends down. The pain limits her activity. Her medical history is significant for osteoarthritis, for which she uses nonsteroidal anti-inflammatory drugs (NSAIDs) for the past two years. She is living with her husband and has 3 children. She doesn't smoke but drinks alcohol occasionally. Her vital signs are normal. On physical examination, there is a small effusion in the right knee. The effusion grew a little larger and she developed a tender swelling in the popliteal fossa and calf. Both the pain and swelling worsened as she bent and straightened her knee.

\noindent\textbf{Patient language:}
I'm a 47-year-old woman, married with 3 kids. I don't smoke and I drink occasionally. I went to the doctor because of pain in my calf and knee when I was bending down. This has been limiting my daily activities. I have been diagnosed with osteoarthritis for which I have taken anti-inflammatory drugs for the past 2 years. The doctor saw a small fluid buildup in my right knee. This buildup became a bit bigger and I have a swollen calf. The pain is worse when I bend and straighten my knee.

\section{Qwen-14B prompted for explanations}
\label{sec:template_justifs}

\subsection{Patient \#21}

\begin{figure}[ht]
    \centering
    \includegraphics[width=\columnwidth]{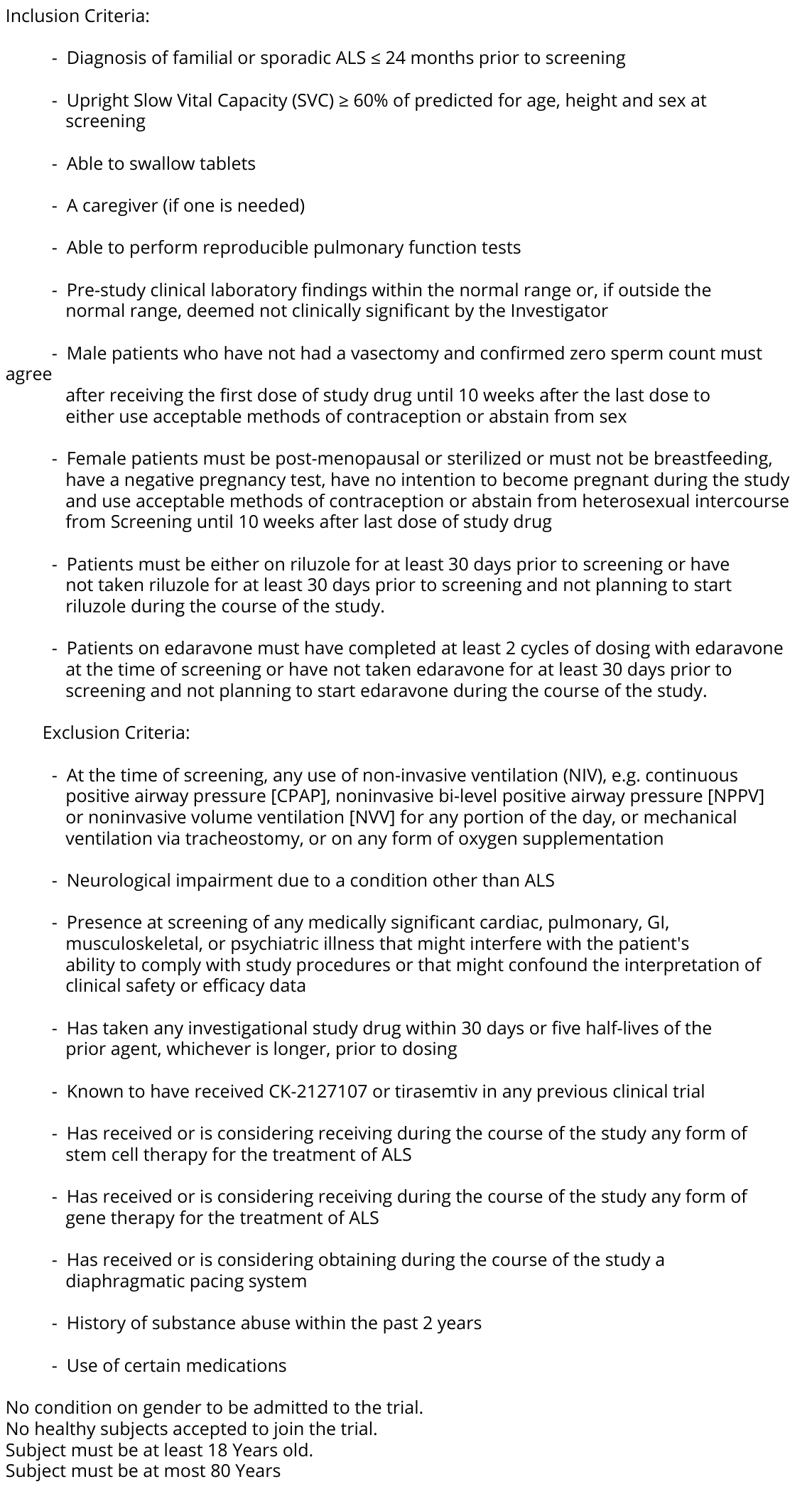}
    \caption{Eligibility criteria from trial NCT03160898, used as the premise.}
    \label{fig:NCT03160898}
\end{figure}

\textbf{Premise (NCT03160898):} See Fig~\ref{fig:NCT03160898}.

\noindent\textbf{Medical answer:}

``Entailment

Explanation: Based on the information provided about the 47-year-old male patient, it seems he meets some key inclusion criteria such as being diagnosed with ALS (within 24 months before screening), currently taking Riluzole which aligns with the pre-screening requirement of having been on riluzole for at least 30 days prior to screening, and being within the specified age range (at least 18 but at most 80 years). However, there isn't enough information given to determine whether his SVC is $\ge$ 60\%, if he has an able caregiver, can swallow tablets, etc., nor is there sufficient detail to ascertain if he fulfills all exclusion criteria (such as absence of neurological impairment not related to ALS, no history of substance abuse within the past two years, specific medical conditions that could interfere, or recent participation in another clinical trial).
The presence of mild gait ataxia mentioned does raise concern potentially regarding the exclusion criterion of "Presence at screening of any medically significant...psychiatric illness" or possibly indicating another neurological disorder; however, without explicit confirmation that this is not due to ALS, we cannot definitively rule him out based solely on this symptom alone. Similarly, without more specifics on the rest of the criteria, particularly those relating to functional capacity and absence of contraindicating health conditions, it is prudent to default towards entailment barring concrete contradictions. Therefore, the answer leans toward entailment pending verification of additional details necessary for full eligibility assessment.''

\noindent\textbf{PL answer:}

``Contradiction

The provided patient information does not specify whether they have been diagnosed with Amyotrophic Lateral Sclerosis (ALS). Furthermore, it mentions that the patient is taking medication for sclerosis rather than riluzole which is typically used for ALS. Additionally, there is no mention of their SVC percentage which is required to meet the inclusion criteria. Hence based solely on the provided information, we cannot conclude that this patient meets all the necessary conditions to be eligible for the trial as described.''

\subsection{Patient \#30}

\begin{figure}[ht]
    \centering
    \includegraphics[width=\columnwidth]{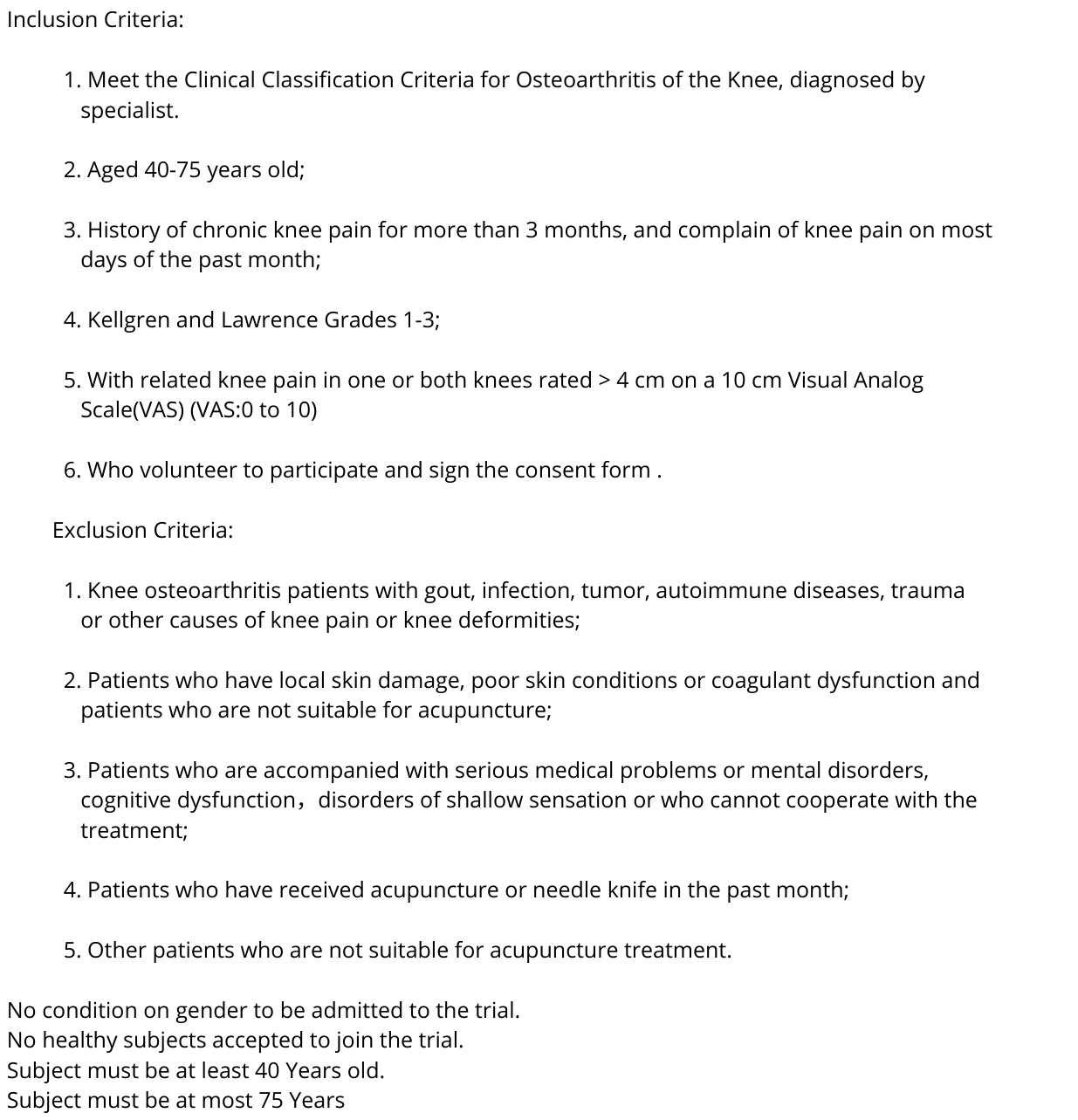}
    \caption{Eligibility criteria from trial NCT03563690, used as the premise.}
    \label{fig:premise_patient_30}
\end{figure}

\textbf{Premise (NCT03563690):} See Fig~\ref{fig:premise_patient_30}.

\noindent\textbf{Medical answer:}

``Contradiction

The patient's presentation includes a tender swelling in the popliteal fossa and calf along with an enlarged effusion due to bending and straightening the knee, which may indicate other causes of her knee pain beyond osteoarthritis, such as possible vascular issues (e.g., deep vein thrombosis). This could fall under the exclusion criteria point number 1 ("Knee osteoarthritis patients with ... other causes of knee pain..."). Therefore, based on this information, it contradicts the full eligibility to participate merely under the inclusion criteria for osteoarthritis without further clarification or exclusion of other potential diagnoses.''

\noindent\textbf{PL Answer:}

``Entailment

Based on the information provided about your condition and comparing it against the inclusion and exclusion criteria outlined in the clinical trial requirements, there is no direct contradiction found that would exclude you from participating in the trial. However, final determination should involve a thorough review by the healthcare professionals involved in the study as they will need to confirm all specific details such as severity grade, pain level on VAS, and absence of any other listed exclusions like concurrent illnesses that could disqualify participation. Given the description of meeting age range, having chronic knee pain due to confirmed osteoarthritis diagnosis, experiencing significant pain which impacts daily life, and willingness to take part (assuming voluntary agreement and signed consent), you appear to meet several key criteria which allows for potential entailment into the trial framework.''

\end{document}